\documentclass[10pt,twocolumn,letterpaper]{article}

\usepackage{cvpr}
\usepackage{times}
\usepackage{epsfig}
\usepackage{graphicx}
\usepackage{amsmath}
\usepackage{amssymb}
\usepackage{adjustbox}
\usepackage{subcaption}
\usepackage{caption}
\usepackage{nicefrac}
\usepackage{dblfloatfix}
\usepackage{mathrsfs}
\usepackage{xcolor}
\usepackage{colortbl}
\usepackage{amsfonts}

\usepackage{pifont}% http://ctan.org/pkg/pifont
\newcommand{\cmark}{\ding{51}}%
\newcommand{\xmark}{\ding{55}}%

\newcommand\blfootnote[1]{%
  \begingroup
  \renewcommand\thefootnote{}\footnote{#1}%
  \addtocounter{footnote}{-1}%
  \endgroup
}

\definecolor{sh_gray}{rgb}{0.84,0.84,0.84}
\definecolor{sh_gray2}{rgb}{1,0.89,0.75}
\definecolor{color3}{rgb}{0.95,0.95,0.95}
\definecolor{color4}{rgb}{0.96,0.96,0.86}
\definecolor{color5}{rgb}{0.90,0.90,0.90}

\newlength{\Oldarrayrulewidth}

\usepackage[pagebackref=true,breaklinks=true,letterpaper=true,colorlinks,bookmarks=false]{hyperref}

\cvprfinalcopy % *** Uncomment this line for the final submission

\def\cvprPaperID{13} % *** Enter the cvpr Paper ID here

\ifcvprfinal\fi
\begin{document}

%%%%%%%%% TITLE
\title{iSAID: A Large-scale Dataset for Instance Segmentation in Aerial Images}

\author{Syed Waqas Zamir$^{1,*}$ \quad Aditya Arora$^{1,*}$ \quad Akshita Gupta$^{1}$ \quad Salman Khan$^{1}$ \quad Guolei Sun$^{1}$ \\ Fahad Shahbaz Khan$^{1}$ \quad Fan Zhu$^{1}$ \quad Ling Shao$^{1}$ \quad 
Gui-Song Xia$^{2}$ \quad Xiang Bai$^{3}$ \quad \\
$^{1}$Inception Institute of Artificial Intelligence, UAE, 
$^{2}$Wuhan University, China\\
$^{3}$Huazhong University of Science and Technology, China \\
{\tt\small $^{1}${firstname.lastname}@inceptioniai.org}\\
{\tt\small $^{2}$guisong.xia@whu.edu.cn}, 
{\tt\small $^{3}$xbai@hust.edu.cn}
}

\maketitle
%\thispagestyle{empty}

%%%%%%%%% ABSTRACT
\begin{abstract}
\blfootnote{$^*$Equal contribution}
Existing Earth Vision datasets are either suitable for semantic segmentation or object detection. In this work, we introduce the first benchmark dataset for instance segmentation in aerial imagery that combines instance-level object detection and pixel-level segmentation tasks. In comparison to instance segmentation in natural scenes, aerial images present unique challenges e.g., a huge number of instances per image, large object-scale variations and abundant tiny objects. Our large-scale and densely annotated Instance Segmentation in Aerial Images Dataset (iSAID) comes with 655,451 object instances for 15 categories across 2,806 high-resolution images. Such precise per-pixel annotations for each instance ensure accurate localization that is essential for detailed scene analysis. Compared to existing small-scale aerial image based instance segmentation datasets, iSAID contains 15$\times$ the number of object categories and 5$\times$ the number of instances. We benchmark our dataset using two popular instance segmentation approaches for natural images, namely  Mask R-CNN and PANet.  In our experiments we show that direct application of off-the-shelf Mask R-CNN and PANet on aerial images provide suboptimal instance segmentation results, thus requiring specialized solutions from the research community. The dataset is publicly available at: {\small\url{https://captain-whu.github.io/iSAID/index.html}}
\end{abstract}

%%%%%%%%% BODY TEXT
\section{Introduction}

\begin{figure}[t] %fig1
\begin{center}
    \begin{subfigure}[t]{0.15\textwidth}
      \includegraphics[width=\linewidth]{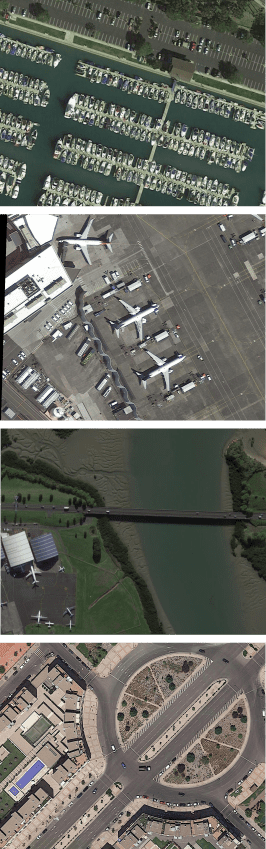}
      \caption{Original image}
      \label{}
    \end{subfigure}
    \begin{subfigure}[t]{0.15\textwidth}
      \includegraphics[width=\linewidth]{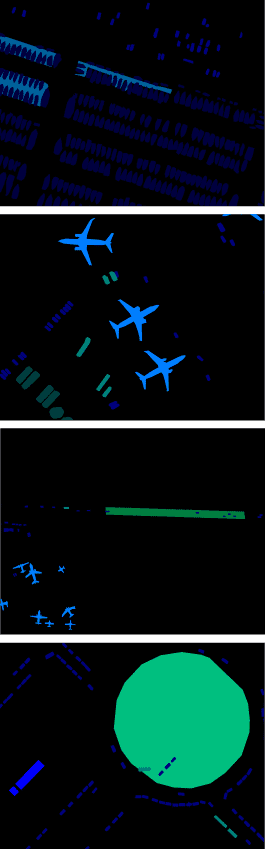}
      \caption{SS maps}
      \label{}
    \end{subfigure}
    \begin{subfigure}[t]{0.15\textwidth}
      \includegraphics[width=\linewidth]{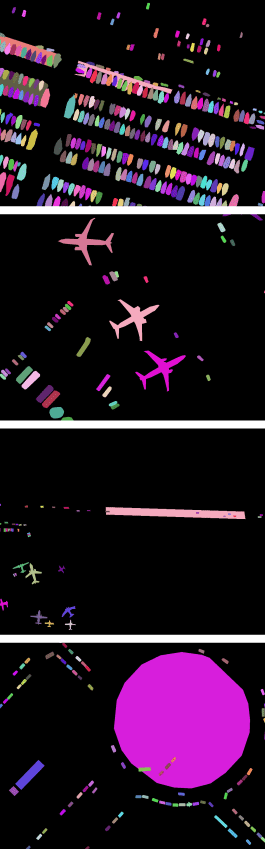}
      \caption{IS maps}
      \label{}
    \end{subfigure}
\end{center}\vspace{-1em}
\caption{Some typical examples from iSAID containing objects with high density, arbitrary shapes and orientation, large aspect ratios and huge scale variation. SS and IS denote semantic segmentation and instance segmentation, respectively. %\textcolor{red}{fig1}
}
\label{fig:teaser}\vspace{0cm}
\end{figure}

Given an image, the aim of instance segmentation is to predict category labels of all objects of interest and localize them using pixel-level masks. 
Large-scale datasets such as ImageNet \cite{imagenet}, PASCAL-VOC \cite{pascal}, MSCOCO \cite{mscoco},  Cityscapes \cite{cityscapes} and ADE20K \cite{ade20k} contain natural scenes in which objects appear with upward orientation. These datasets enabled deep convolutional neural networks (CNN), that are data hungry in nature \cite{khan2018guide}, to show unprecedented performance in scene understanding tasks such as image classification \cite{Simonyan2014,He2016,szegedy2017inception}, object detection \cite{ren2015faster,redmon2017yolo9000,liu2016ssd}, semantic labeling and instance segmentation \cite{he2017mask,liu2018path,chen2018encoder}. 
However, the algorithms developed to solve these tasks in regular images do not transfer well to overhead (aerial) imagery. In aerial images, objects occur in high density (Fig.~\ref{fig:teaser}, row 1), arbitrary shapes and orientation (Fig.~\ref{fig:teaser}, row 2), large aspect ratios (Fig.~\ref{fig:teaser}, row 3), and with huge scale variation (Fig.~\ref{fig:teaser}, row 4). To accurately address the challenges of aerial images for high-level vision tasks, tailor-made solutions on appropriate datasets are desired.  

To encourage the advancements in aerial imagery for earth observation, a few well-annotated datasets for object detection \cite{xview,dota} and semantic labeling \cite{maggiori2017dataset, Urban3D2017} have recently been introduced. However, they do not provide per-pixel accurate labelings for each object instance in an aerial image and are therefore unsuitable for instance segmentation task (see Table \ref{Table:compare datasets}). Publicly available instance segmentation datasets \cite{airbus,spacenet} typically focus on a single object category; for example, \cite{spacenet} only contains building footprints and \cite{airbus} only has labelings for ships. To address the shortcomings of these existing datasets, we introduce a large-scale Instance Segmentation in Aerial Images Dataset (iSAID). Our dataset contains annotations for an enormous 655,451 instances of 15 categories in 2,806 high-resolution images. Having such large number of instances and class count makes iSAID suitable for real-world applications in complicated aerial scenes. 

Compared to other aerial datasets for instance segmentation, iSAID is far more diverse, comprehensive and challenging. It exhibits the following distinctive characteristics: \textbf{(a)} large number of images with high spatial resolution, \textbf{(b)} fifteen important and commonly occurring categories, \textbf{(c)} large number of instances per category, \textbf{(d)} large count of labelled instances per image, which might help in learning contextual information, \textbf{(e)} huge object scale variation, containing small, medium and large objects, often within the same image, \textbf{(f)} Imbalanced and uneven distribution of objects with varying orientation within images, depicting real-life aerial conditions, \textbf{(g)} several small size objects, with ambiguous appearance, can only be resolved with contextual reasoning, \textbf{(h)} precise instance-level annotations carried out by professional annotators, cross-checked and validated by expert annotators complying with well-defined guidelines.

\section{Related Work}
Both in terms of historical context and recent times, large-scale datasets have played a key role in progressing the state-of-the-art for scene understanding tasks such as image classification \cite{Simonyan2014,He2016,szegedy2017inception}, scene recognition \cite{places}, object detection \cite{ren2015faster,redmon2017yolo9000,liu2016ssd} and segmentation \cite{he2017mask,liu2018path,chen2018encoder}. For instance, ImageNet \cite{imagenet} is one of the most popular large-scale dataset for image classification task, on which the state-of-the-art methods \cite{Simonyan2014,He2016,szegedy2017inception} are able to reach human-level performance. 
Similarly, large-scale annotated datasets, such as MSCOCO \cite{mscoco},  Cityscapes \cite{cityscapes} and ADE20K \cite{ade20k} for object detection, semantic and instance segmentation have driven the development of exciting new solutions for natural scenes.
%Similar trends have been observed on other conventional imagery datasets: PASCAL VOC \cite{pascal}, MSCOCO \cite{mscoco} for object detection and instance segmentation, Cityscapes \cite{cityscapes} and ADE20K \cite{ade20k} for semantic segmentation, and Places \cite{places} for scene recognition.
Introduction of datasets, that are larger in scale and diversity, not only provide room for new applications but also set new research directions. Moreover, challenging datasets push research community to develop more sophisticated and robust algorithms; thus enabling their application in real-world scenarios.  

There are numerous lucrative application areas of Earth Vision research, including security and surveillance \cite{santamaria2017mass}, urban planning \cite{marcos2018land}, %agriculture field monitoring, 
precision agriculture, land type classification \cite{Anwer2018} and change detection \cite{khan2017forest}. In general, deep-learning based algorithms show excellent performance when provided with large-scale datasets, as demonstrated for several high-level vision tasks \cite{He2016,he2017mask,redmon2017yolo9000} involving conventional large-scale image datasets. A key limitation towards building solutions for the Earth Vision applications is the unavailability of aerial datasets resembling the scale and diversity of natural-scene datasets (\eg ImageNet \cite{imagenet} and MSCOCO \cite{mscoco}). Specifically, existing overhead imagery datasets are significantly lagging in terms of category count, instance count and the quality of annotations.  
The advanced off-the-shelf methods trained on conventional datasets when applied on aerial image datasets, fail to provide satisfactory results due to large domain shift, high density objects with large variations in orientation and scale. As an example, an otherwise robust object detector SSD \cite{liu2016ssd} yields an mAP of just 17.84 on the dataset for object detection in aerial images (DOTA) \cite{dota}. 
Recently, large-scale aerial image datasets (DOTA \cite{dota} and xView \cite{xview}) have been introduced to make advancement in object detection research for earth observation and remote sensing.  Both of these datasets \cite{dota,xview} are more diverse, complex, and suitable for real-world applications than previously existing aerial datasets for object detection \cite{nwpu,sztaki,cowc,ucas,hrsc}. On the down side, these datasets do not provide pixel-level masks for the annotated object instances. 

 Instance segmentation is a challenging problem that goes one step ahead than regular object detection as it aims to achieve precise per-pixel localization for each object instance. Unlike aerial object detection, there exist no large-scale annotated dataset for instance segmentation in aerial images. A few publicly available datasets in this domain only contain instances of just a single category (e.g., ships \cite{airbus} and buildings \cite{spacenet}). Owing to the significance of precise localization of each instance in aerial imagery, we introduce a novel dataset, iSAID, that is significantly large, challenging, well-annotated, and offers 15$\times$ the number of object categories and 5$\times$ the number of instances than existing datasets \cite{airbus,spacenet}.

\section{Dataset Details}
\subsection{Images, Classes and Dataset Splits}
\begin{figure}[t] %fig6
\begin{center}
    \begin{subfigure}[t]{0.495\columnwidth}
      \includegraphics[width=\linewidth]{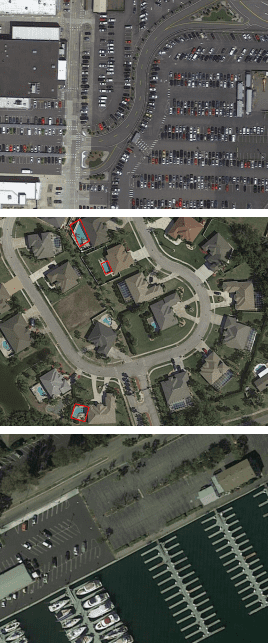}
      \caption{DOTA}
      \label{}
    \end{subfigure}
    \begin{subfigure}[t]{0.495\columnwidth}
      \includegraphics[width=\linewidth]{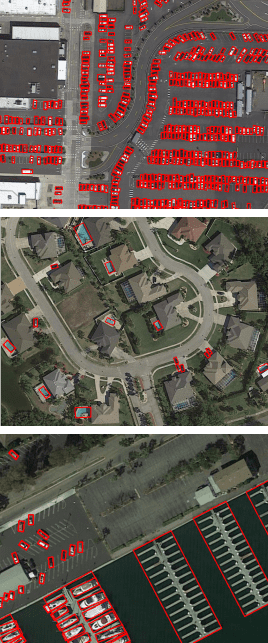}
      \caption{iSAID}
      \label{}
    \end{subfigure}
\end{center}\vspace{-1.5em}
\caption{Visualization of missing annotations from DOTA~\cite{dota} as compared to iSAID. %\textcolor{red}{fig6}
}
\label{fig:missing_vis}\vspace{0cm}
\end{figure}

In order to create a dataset for instance segmentation task, we build on the large-scale aerial image dataset: DOTA \cite{dota}, that contains 2,806 images. The images are collected from multiple sensors and platforms to reduce bias. Note that the original DOTA dataset only contains bounding box annotations for object detection, thus cannot be used for accurate instance segmentation. Furthermore, DOTA \cite{dota} suffers with several aberrations such as incorrect labels, missing instance annotations, and inaccurate bounding boxes. To avoid these issues, our dataset for instance segmentation is \textbf{independently annotated from scratch}, leading to 655,451 instances compared to 188,282 instances provided originally in DOTA \cite{dota} (a $\sim250\%$ relative increase, see Fig.~\ref{fig:missing_vis} for examples). 

\begin{figure}[!t] %fig5
\begin{center}
\includegraphics[width=\linewidth]{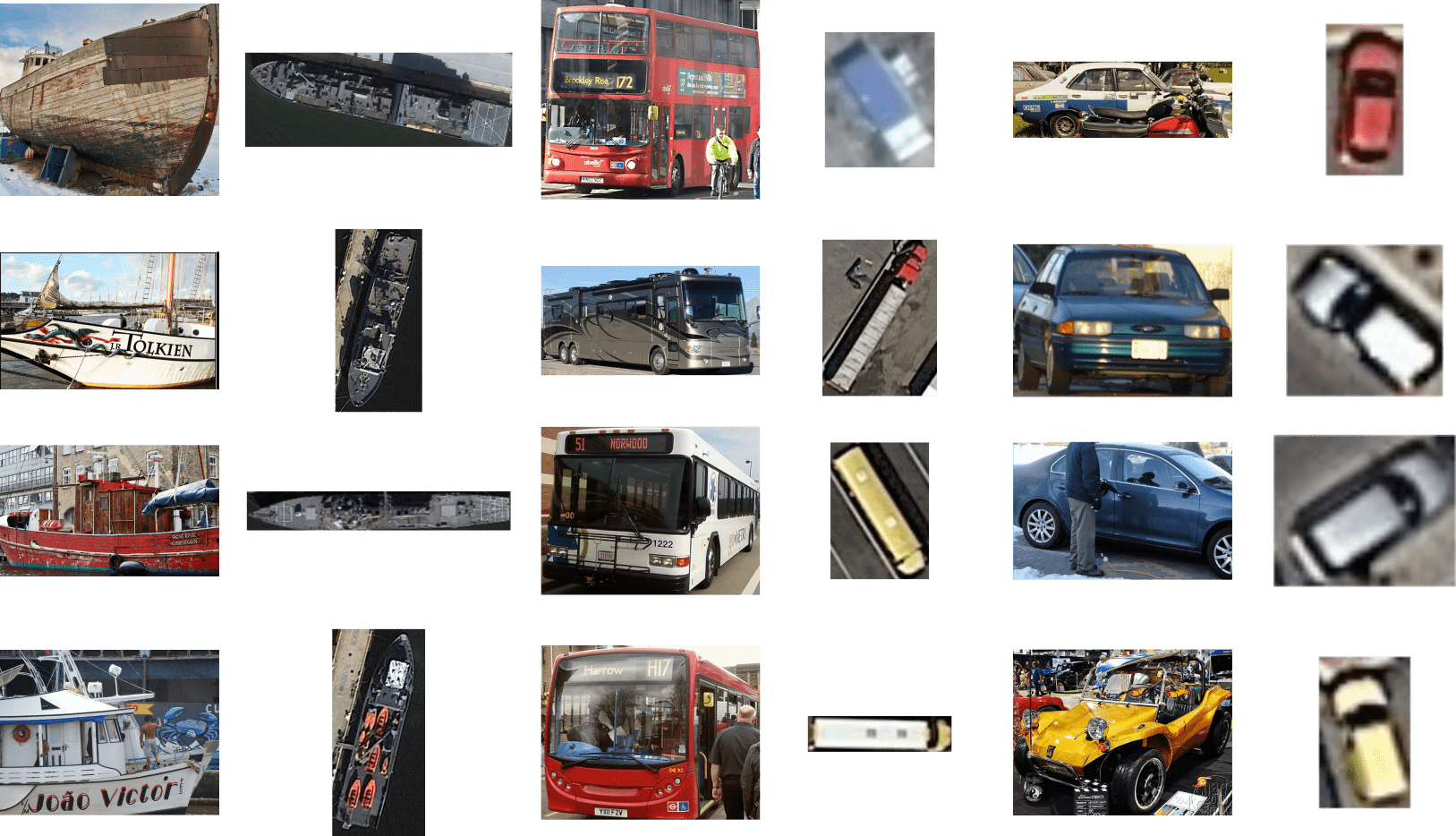}
\end{center}\vspace{-1em}
\caption{Ships, buses and cars from MSCOCO~\cite{mscoco} (odd columns) and iSAID (even columns). Notice the size variation and the angle at which images are taken. %\textcolor{red}{fig5}
}
\label{fig:compCOCO}\vspace{0cm}
\end{figure}

\begin{figure}[t] %fig4
\centering
\includegraphics[width = 1\linewidth]{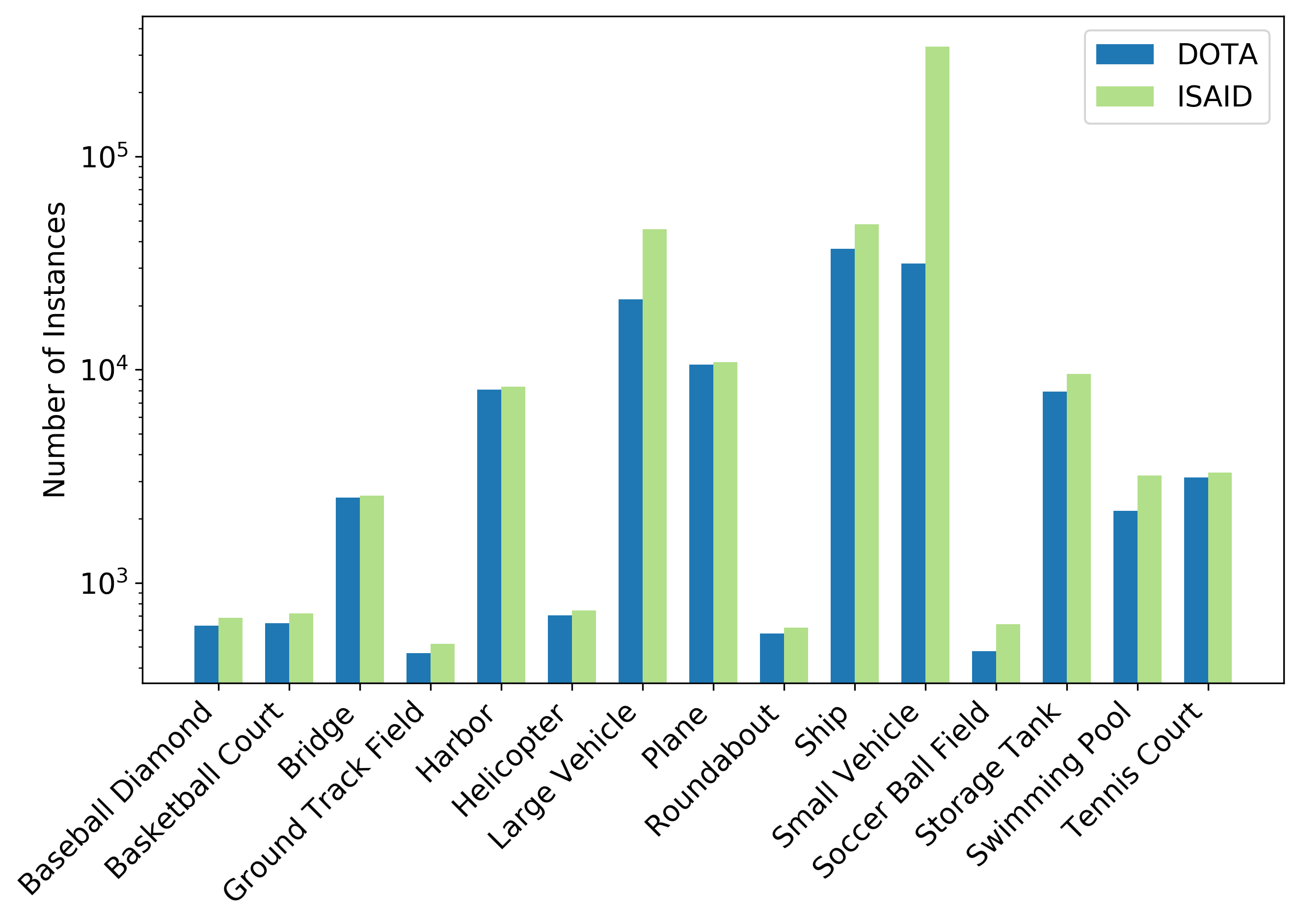} \vspace{-0.4cm} 
\caption{Comparison of DOTA \cite{dota} and our dataset (iSAID) in terms of instances per category. iSAID contains, in total, 3.5 times more number of instances than DOTA. %\textcolor{red}{fig4}
} 
\label{Fig:dota_classwise_comp} 
\end{figure}

It is important to note that the our instance segmentation dataset in aerial images has unique challenges compared to regular image datasets (e.g., less object details, small size and different viewpoints- see Fig.~\ref{fig:compCOCO}).  On the other hand, as summarized in Table~\ref{Table:compare datasets}, most of the existing aerial image datasets are annotated with bounding boxes or point-labels that only coarsely localize the object instances. Furthermore, these datasets are often limited to a small scale with only a few object categories. In comparison, our proposed iSAID dataset provides a large number of instances, precisely marked with masks denoting their exact location in an image (Fig.~\ref{Fig:dota_classwise_comp}). The two existing instance segmentation datasets  for aerial imagery only comprise of a single object category (e.g., ships \cite{airbus} or buildings \cite{spacenet}). In contrast, iSAID has a diverse range of 15 categories and much larger scale ($\sim$5$\times$ more instances).

\begin{table*}[!htp]
    \centering
     \adjustbox{width=1\textwidth}{%
    \setlength{\tabcolsep}{5pt}
    \begin{tabular}{c|c c c c c c c c}
        \hline
        Dataset & Bounding & Segmentation &  \#Main &  \#Fine-grain & \#Total  & \#Instances & \#Images  & Image \\
        & box & mask &  categories &  categories & categories &  &  & width \\
        \hline
        NWPU VHR-10 \cite{nwpu} & horizontal & \xmark &10 & \xmark & 10& 3,651 & 800 & $\sim$1,000 \\
        SZTAKI-INRIA \cite{sztaki} &  oriented & \xmark & 1 & \xmark & 1 & 665 & 9 & $\sim$800 \\
        TAS \cite{tas} & horizontal & \xmark&  1 & \xmark & 1 & 1,319 &  30 &  792 \\
        COWC \cite{cowc} &  center-point & \xmark&  1 & \xmark & 1 & 32,716 &  53 &  2,000 $\sim$ 19,000 \\ 
        VEDAI \cite{vedai} &  oriented& \xmark &  3 & \cmark& 9 &  3,700  & 1,200 &  512, 1,024 \\
        UCAS-AOD \cite{ucas} &  horizontal & \xmark&  2 & \xmark & 2 & 6,029  & 910 &  $\sim$1,000 \\
        HRSC2016 \cite{hrsc} &  oriented & \xmark&  1 & \xmark & 1 &  2,976 &  1,061 &  300$\sim$1,500 \\
        xView \cite{xview} & horizontal & \xmark & 16 & \cmark & 60 & 1,000,000 & 1,127  & 700$\sim$4,000 \\ 
        DOTA \cite{dota} &  oriented & \xmark &   14 & \cmark & 15 &  188,282 &   2,806 &   800$\sim$13,000 \\
        \hline
        Airbus Ship \cite{airbus} &  polygon & \cmark&  1 & \xmark & 1 &   131,000 &   192,000 & $\sim$800  \\
        SpaceNet MVOI  \cite{spacenet} & polygon & \cmark  & 1 &  \xmark & 1 &  126,747 & 60,000 & 900 \\
        iSAID (Ours) & polygon & \cmark & 14 & \cmark & 15  & 655,451 & 2,806 & 800$\sim$13,000 \\
        \hline
    \end{tabular}}
    \caption{Comparison between Aerial Datasets. $\emph{Center-point}$ represents those annotations for which only the center coordinates of the instances are provided. 
    }
    \label{Table:compare datasets}
\end{table*}

In order to select object categories we follow the experts in overhead satellite imagery interpretation \cite{dota} and provide annotations for the following 15 classes: \emph{plane, ship, storage tank, baseball diamond, tennis court, basketball court, ground track field, harbor, bridge, large vehicle, small vehicle, helicopter, roundabout, swimming pool} and  \emph{soccer ball field}. Objects from these categories occur frequently and are important for various real-world applications \cite{nwpu,cowc,ucas}. For dataset splits, we use half of the original images to form train set, $1/6$ images for validation set and $1/3$ for test set. Both images and ground-truth annotations for the train and validation sets will be released publicly. In the case of test set, we will publicly provide images without annotations. The test set annotations will be used to set up an evaluation server for fair comparison between the developed techniques.

%\SK{One para about "iSAID vs Other Datasets" with reference to compairson table and figures. }

\begin{figure*}[t]
\begin{center}
    \begin{subfigure}[t]{0.24\textwidth} %fig2
      \includegraphics[width=\linewidth]{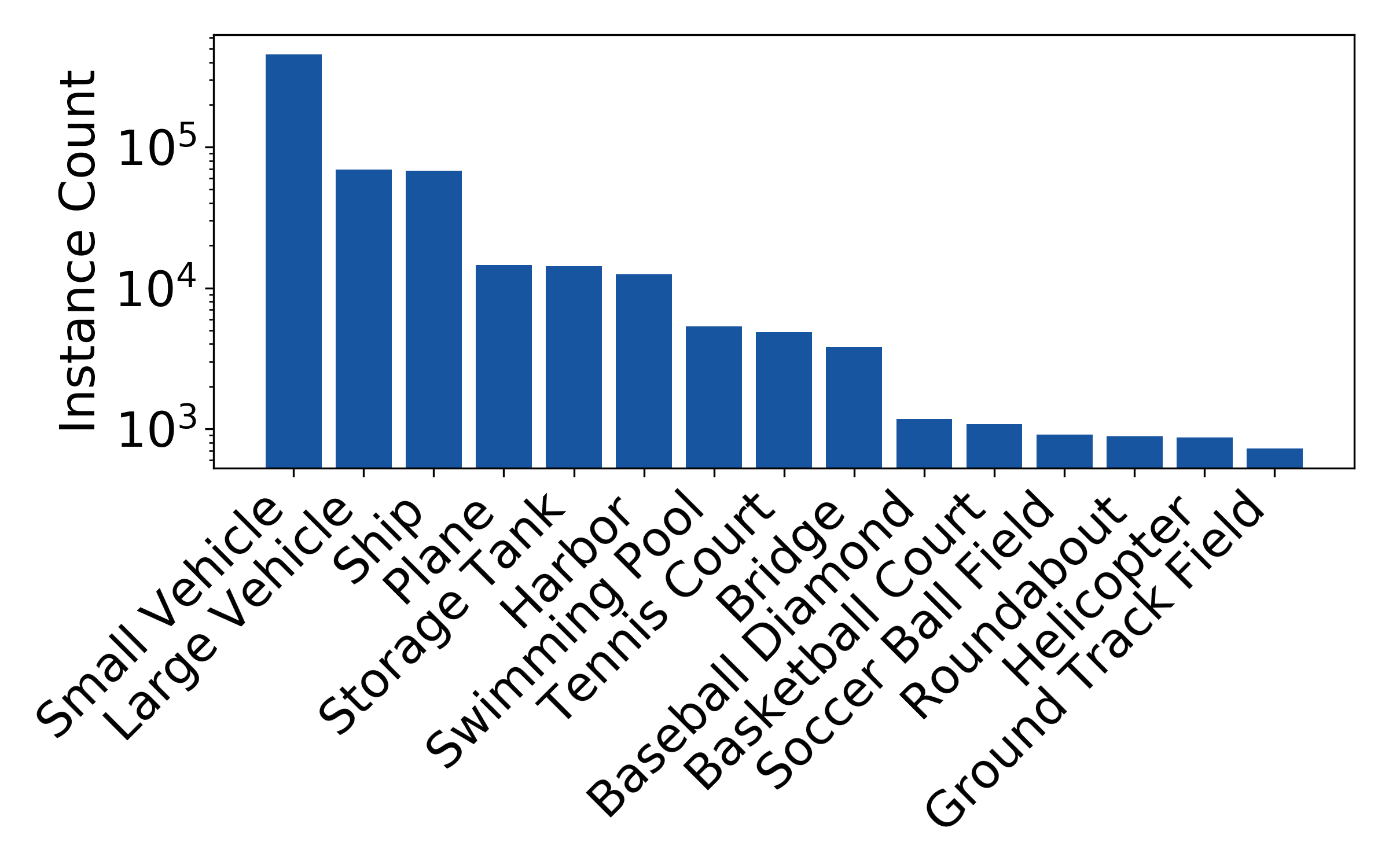}
       \caption{}%{\textcolor{red}{fig2}}
      \label{fig:instances_per_category}
    \end{subfigure}
    \begin{subfigure}[t]{0.24\textwidth} %fig81
      \includegraphics[width=\linewidth]{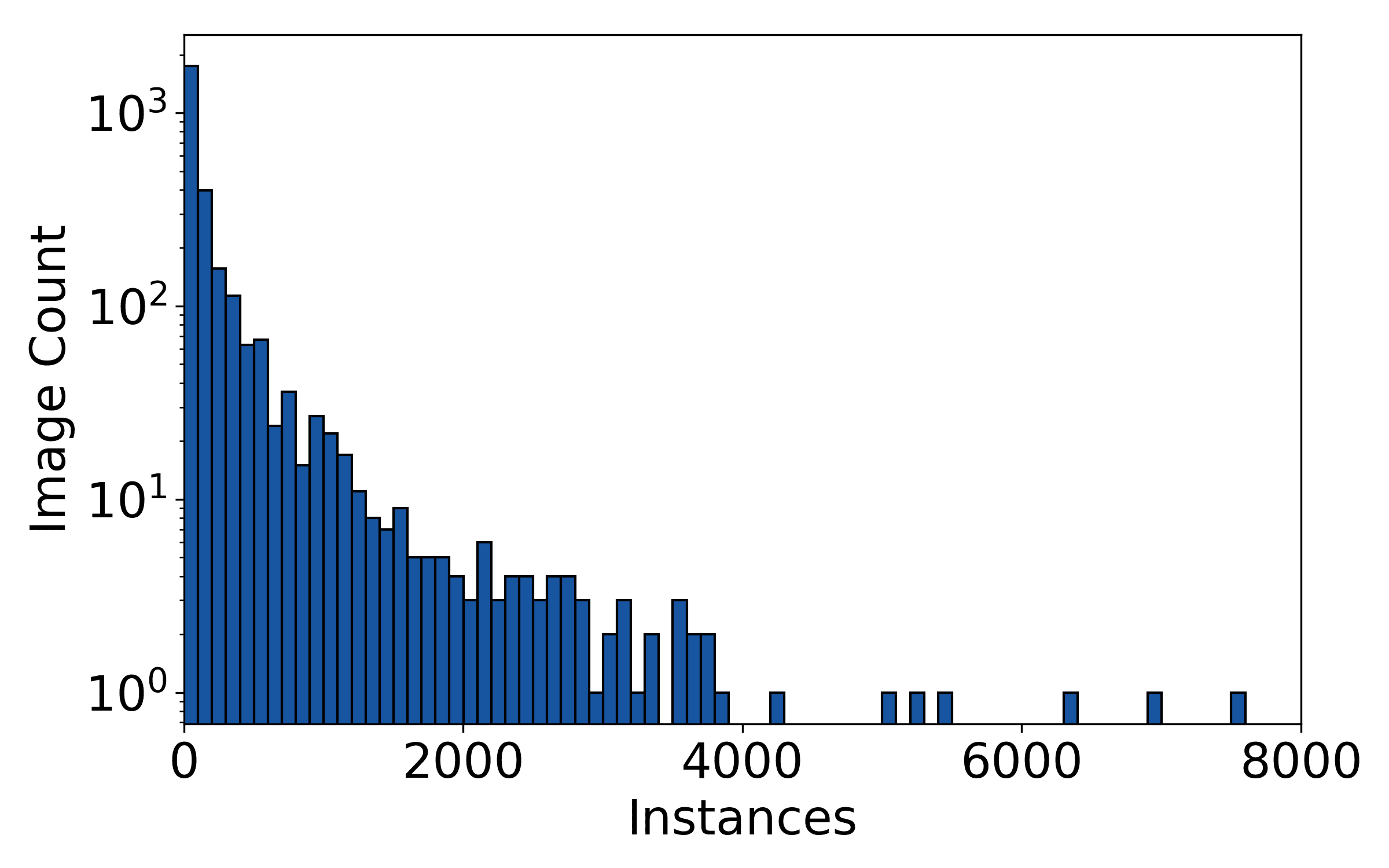}
       \caption{}%{\textcolor{red}{fig81}}
      \label{fig:instances per image}
    \end{subfigure}
    \begin{subfigure}[t]{0.24\textwidth} %fig82
        \includegraphics[width=\linewidth]{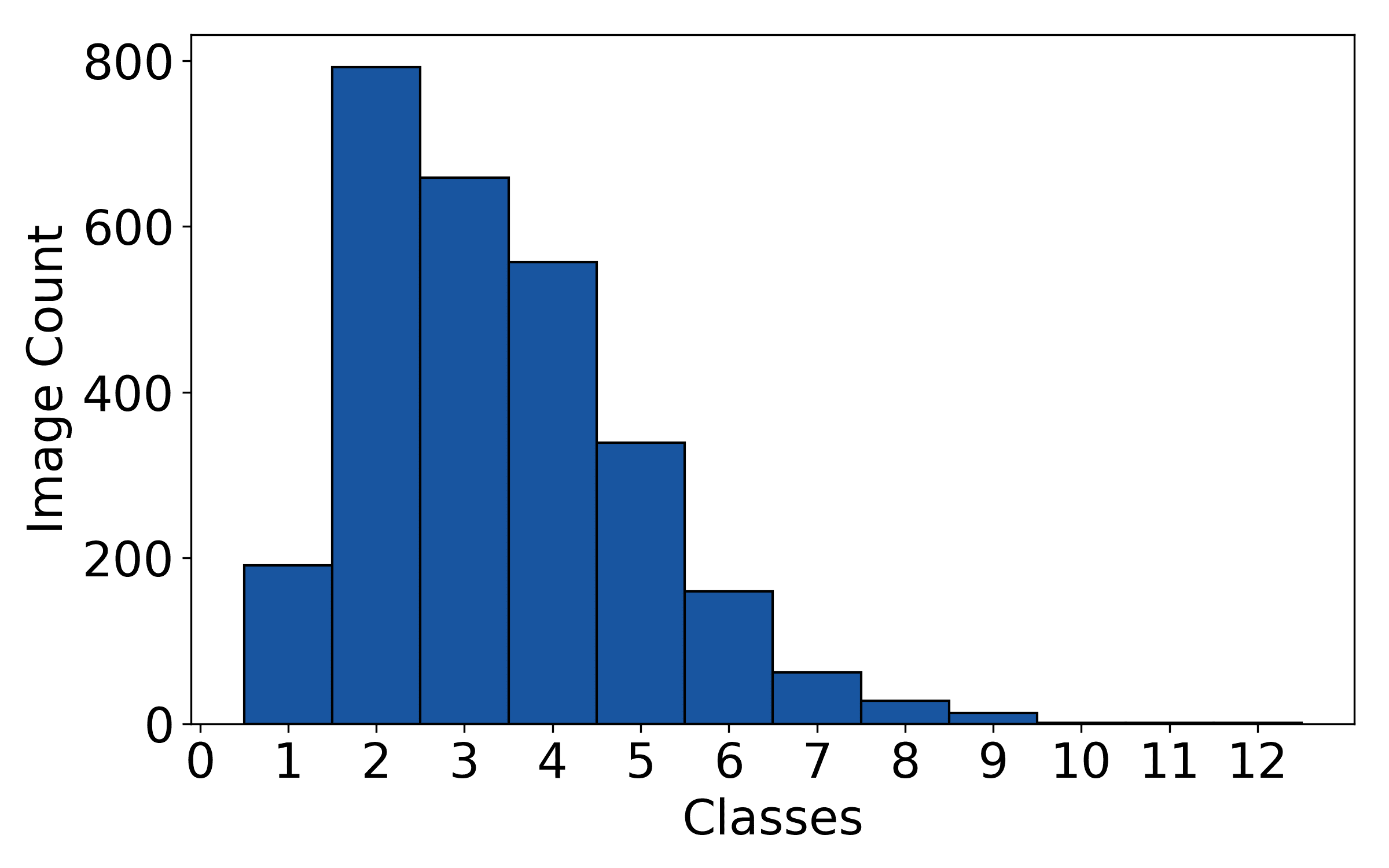}
        \caption{}%{\textcolor{red}{fig82}}
        \label{fig:classes_per_image}\vspace{0cm}
    \end{subfigure}
    \begin{subfigure}[t]{0.24\textwidth} %fig11
      \includegraphics[width=\linewidth]{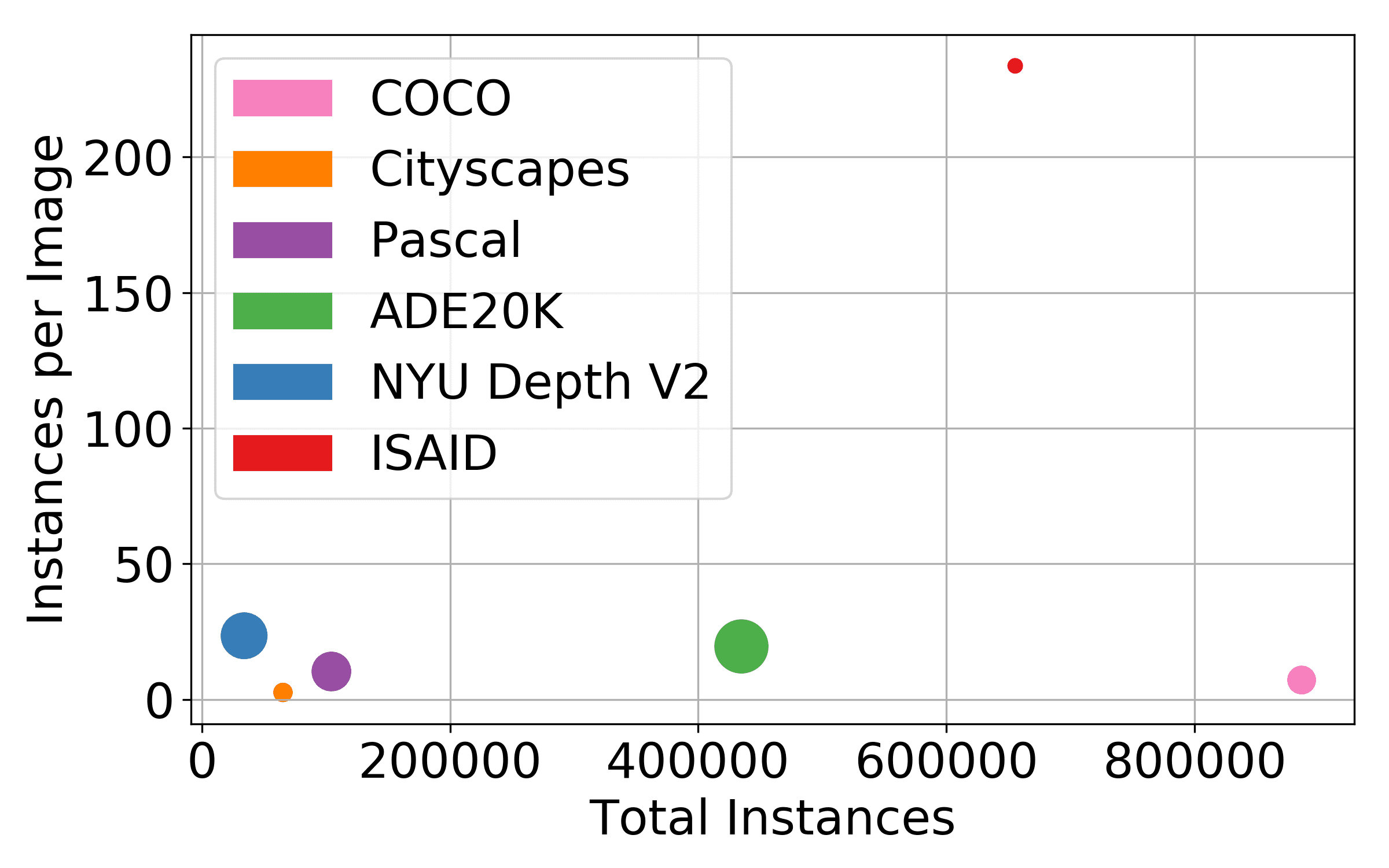}
       \caption{}%{\textcolor{red}{fig11}}
      \label{fig:datasets}
    \end{subfigure}
\end{center}\vspace{-1.5em}
\caption{Statistics of classes and instances in iSAID. (a) Histogram of the number of instances per class (sorted by frequency). (b) Histogram of number of instances per image. (c) Histogram of number of classes per image. (d) Number of instances vs. instances per image (comparison of our dataset with other large-scale conventional datasets). The size of the circle denotes the number of categories, \eg, big circle represents the presence of large number of object categories.}
\label{fig:faster_rcnn}\vspace{-0cm}
\end{figure*}

\subsection{Annotation Procedure}
We design a comprehensive annotation pipeline to ensure that annotations of all images are consistent, accurate and complete. The pipeline includes the following steps: developing annotation guidelines; training annotators; annotating images; quality checks and annotation refinement until satisfaction. For annotation, a high-quality in-house software named \emph{Haibei} was used to draw instance segmentation masks on images. 

In order to obtain high-quality annotations, clear and thorough guidelines for annotators are of prime importance. Taking notes from previously proposed datasets \cite{dota,xview,mscoco,ade20k,pascal}, we establish the following guidelines: \textbf{1)} All clearly visible objects of the above-mentioned 15 categories must be annotated; \textbf{2)} Segmentation masks for each instance should match its visual margin in the image; \textbf{3)} Images should be zoomed in or out, when necessary, to obtain annotations with refined boundaries; \textbf{4)} Cases of unclear/difficult objects should be reported to the team supervisors and then discussed to get annotations with high confidence; \textbf{5)} All work should be done at a single facility using the same software.% in order to maintain single annotation standard.

The images of proposed iSAID are annotated by the professional annotators. The annotators were trained through multiple sessions, even if they had prior experience in annotating datasets of any kind.
During training phase, each annotator was shown both positive and negative examples containing objects from 15 categories. An assessment protocol was developed to shortlist the best annotators in the following manner: annotators were asked to annotate several sample images containing easy and difficult cases while strictly adhering to the established guidelines. The quality of annotations was crossed checked to evaluate their performance. Only those annotators who passed the test were approved to work on this particular project. In general, the selected annotators were given training for approximately 4 hours before assigning them the task of annotating actual aerial image dataset.

At the beginning of the annotation process, the supervisory team distributes different sets of images among annotators. The annotators were asked to annotate all objects belonging to 15 categories appearing in the images. Due to high spatial resolution and large number of instances, it took approximately 3.5 hours for one annotator to finish labelling all objects present in a single image, resulting in 409 man-hours  (for 2,806 images) excluding cross checks and refinements.

Once the first round of annotations was completed, a five-stage quality control procedure was put in place to ensure that the annotation quality is good. \textbf{1)} The labelers were asked to examine their own annotated images and correct issues like double labels, false labels, missing objects and inaccurate boundaries. \textbf{2)} The annotators reviewed the work of other peers on rotational basis. In this stage, object masks for each class were cropped and placed in one specific directory, so that the annotation errors could be easily identified and corrected. \textbf{3)} The supervisory team randomly sampled 70\% images (around 2000) and analyzed their quality. \textbf{4)}
%Expert team leads sampled
A team of experts sampled 20\% images (around 500) and ensured the quality of annotations. In case of problems, the annotations were iteratively send back to the annotators for refinement until the experts were satisfied by the labels. %95\% correctness rate was achieved. 
\textbf{5)} Finally, several statistics (\eg, instance areas, aspect ratios, etc.) were computed. Any outliers were double checked to make sure they are indeed valid and correct annotations.

\subsection{iSAID Statistics}
In this section we analyze the properties of iSAID and compare it with other relevant datasets. 

\noindent
\textbf{Image resolution.} Images in natural datasets (\eg, PASCAL-VOC \cite{pascal}, ImageNet \cite{imagenet}) are generally of limited dimensions, often reaching no more than 1000$\times$1000 pixels.  In contrast, aerial images have a very large resolution: for instance the width of some images in COWC \cite{cowc} dataset is up to 19,000 pixels. In our dataset, the spatial resolution of images ranges from $800$ to $13,000$ in width. Applying off-the-shelf conventional object detection and instance segmentation methods on such high-resolution aerial images yield suboptimal results, as we shall see in the experiment section.

\begin{figure} %fig3
\centering
\includegraphics[width = 1\linewidth]{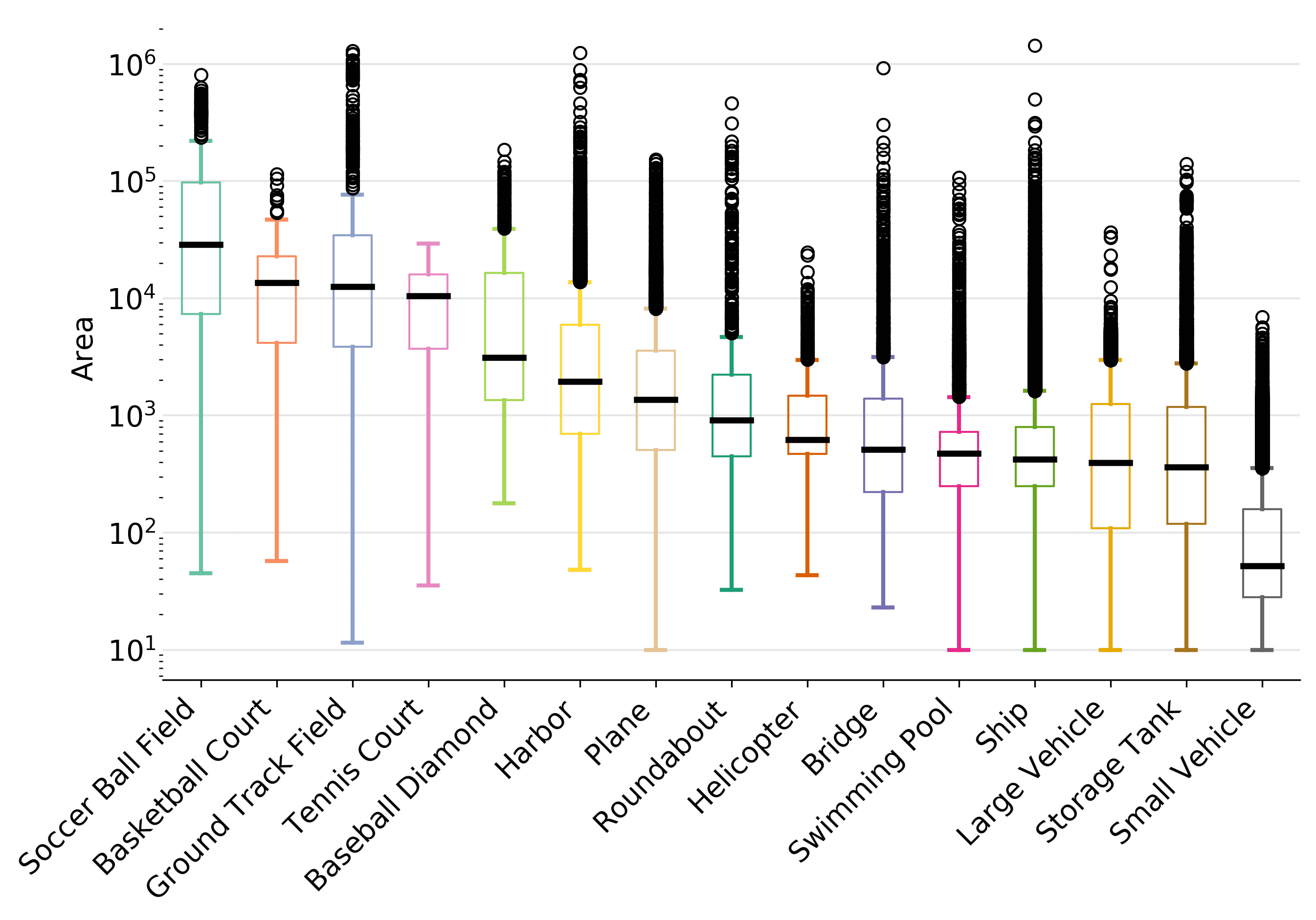} \vspace{-0.4cm}
\caption{Boxplot depicting the range of areas for each object category. The size of objects varies greatly both among and across classes. %\textcolor{red}{fig3}
} 
\label{fig:mean_area_per_category} 
\end{figure}

\begin{figure}[t]
\begin{center}
    \begin{subfigure}[t]{0.49\columnwidth} %fig10
      \includegraphics[width=\linewidth]{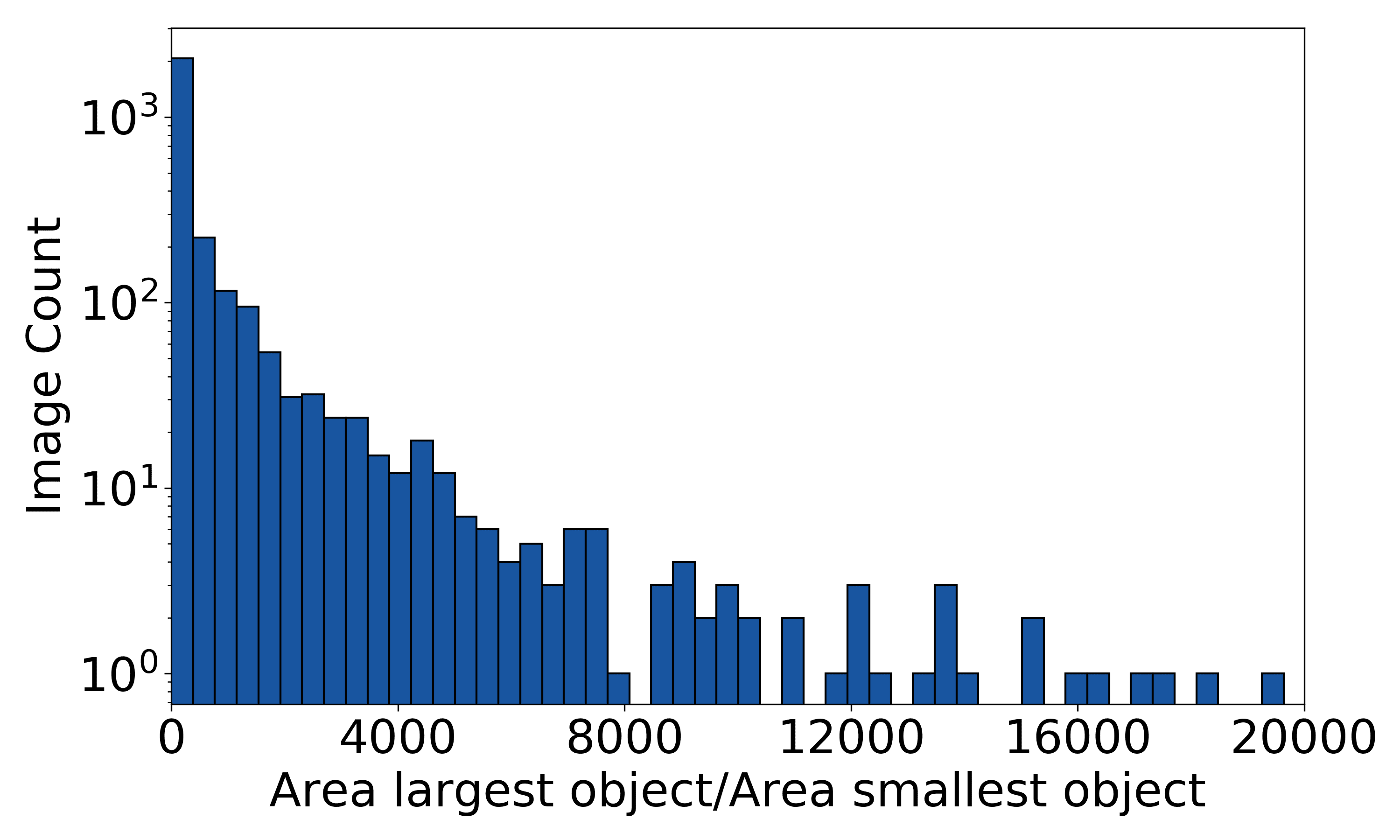}
       \caption{ %\textcolor{red}{fig10}
       }
      \label{fig:scale variation}
    \end{subfigure}
    \begin{subfigure}[t]{0.49\columnwidth} %fig9
      \includegraphics[width=\linewidth]{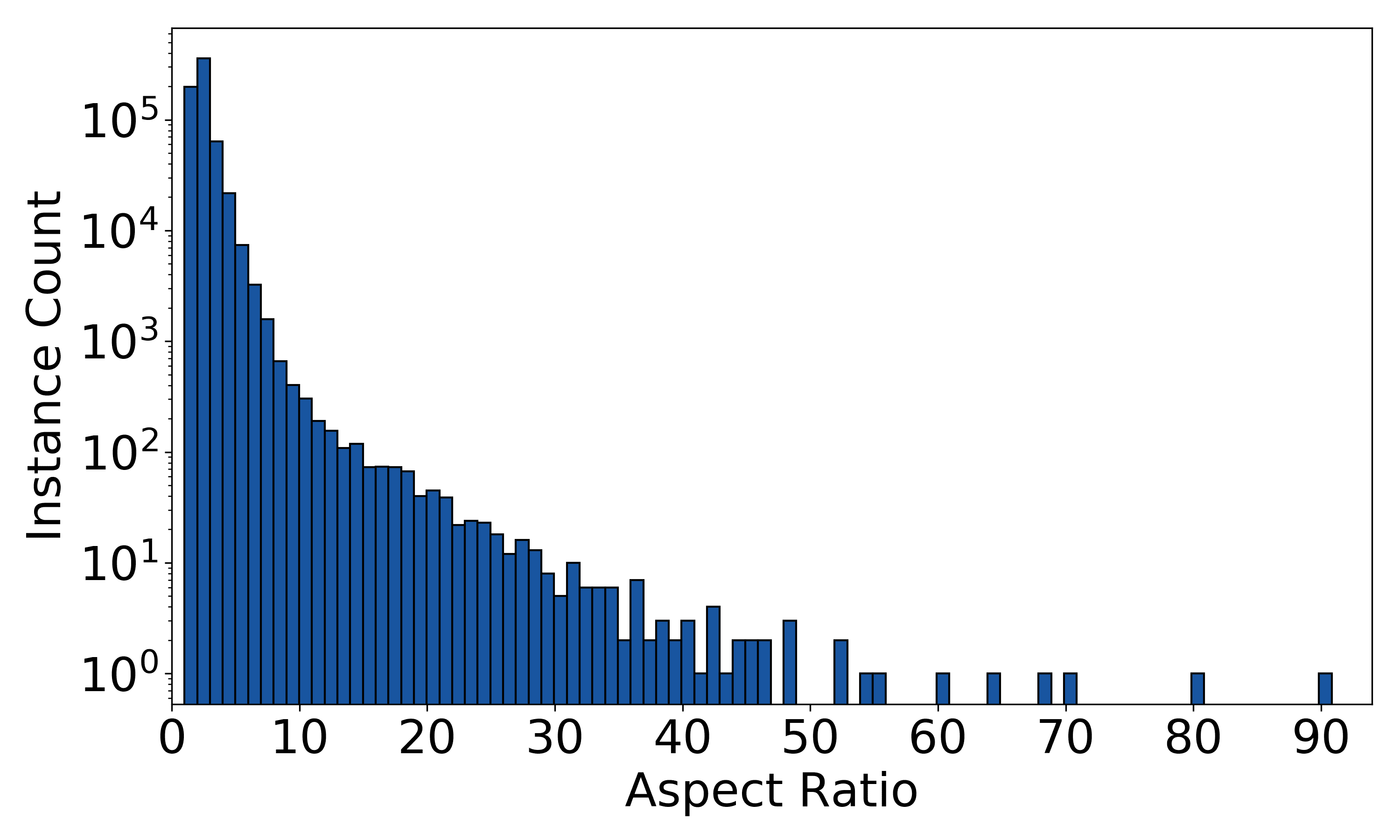}
       \caption{ %\textcolor{red}{fig9}
       }
      \label{fig:aspect ratio}
    \end{subfigure}
\end{center}\vspace{-1.5em}
\caption{Statistics of images and instances in iSAID. (a) Ratio between areas of largest and smallest object shows the huge variation in scale.% when small and large objects of same or different categories appear together. 
(b) shows that instances in iSAID exhibit large variation in aspect ratio.}
\label{fig:faster_rcnn}%\vspace{-0.3cm}
\end{figure}

\begin{figure*}[!htp]
\centering
\includegraphics[width = \linewidth]{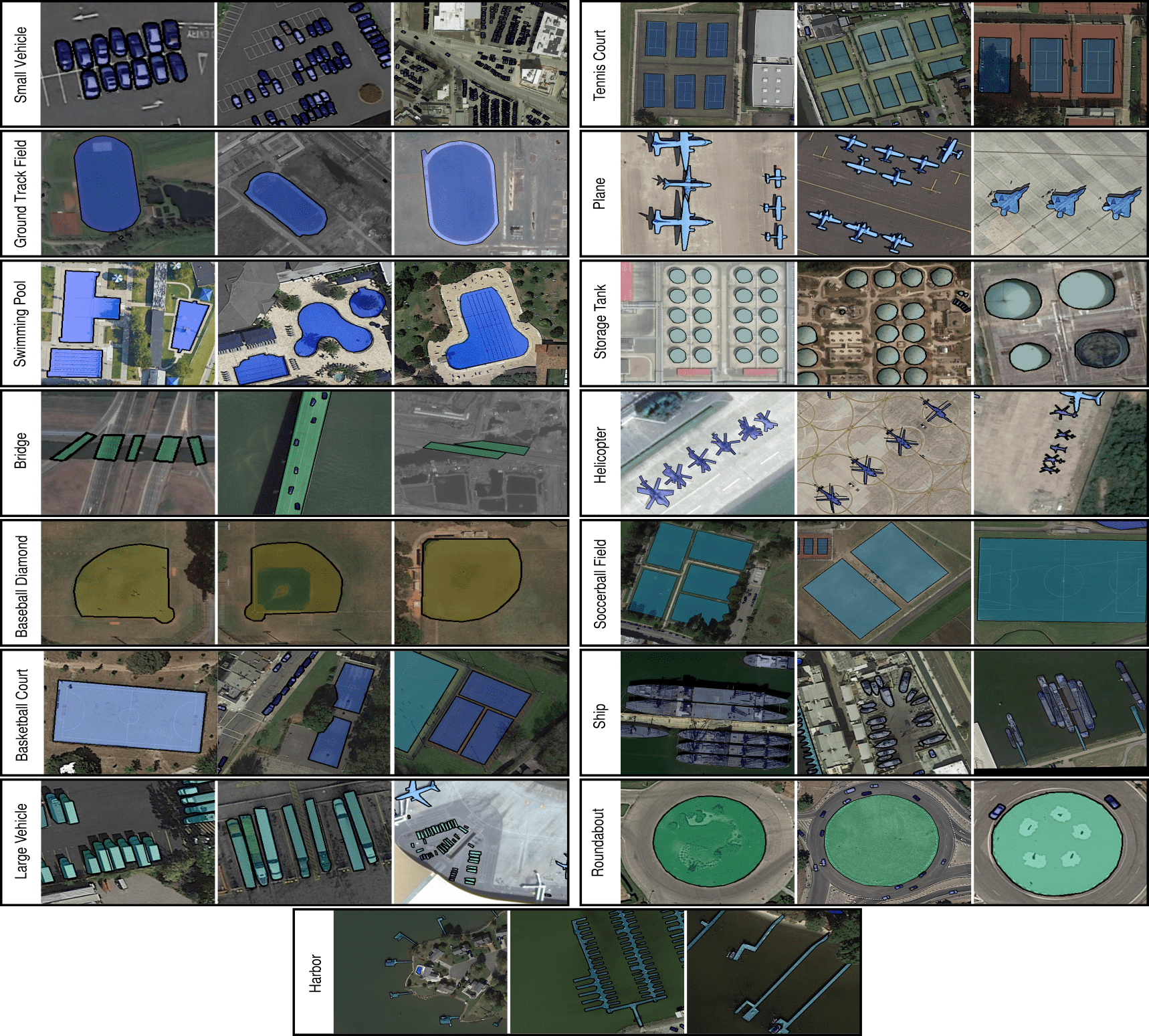}
\caption{Samples of annotated images in iSAID. }
\label{fig:our samples} %\vspace{}
\end{figure*}

\noindent
\textbf{Instance count.} 
Our dataset comprises 655,451 annotated instances of 15 categories. In Fig.~\ref{fig:instances_per_category} it is shown that there are some infrequent classes with significantly less number of instances than other more frequent classes. For example, small vehicle and ground track field are the most frequent and least frequent classes, respectively. Such a class imbalance usually exists in both natural and aerial imagery datasets and it is important for real-world applications \cite{khan2019striking}. Fig.~\ref{fig:classes_per_image} illustrates the image histogram in which multiple classes co-exists; on average 3.27 classes appear in each image of iSAID.

Another property, common in all aerial image datasets, is the presence of large number of object instances per image due to a large field of view. As shown in Fig.~\ref{fig:instances per image}, the instance count per image in our dataset can reach up to 8,000. Fig.~\ref{fig:datasets} depicts that our dataset contains on average $\sim$239 instances per image, which is significantly higher compared to traditional large-scale datasets for instance segmentation: MSCOCO \cite{mscoco}, Cityscapes \cite{cityscapes}, PASCAL-VOC \cite{pascal}, ADE20K \cite{ade20k} and NYU Depth V2 \cite{nyu} contain 7.1, 2.6, 10.3, 19.5, and 23.5 instances per image, respectively. In aerial images, the densely packed instances typically appear in scenes containing parking lots and marina.

\noindent
\textbf{Area of categories.} 
In natural as well as aerial images, objects appear in various sizes. Therefore, an instance segmentation method should be flexible and efficient enough to deal with objects of small, medium and large sizes \cite{dota}. In our dataset, we consider objects in the range 10 to 144 pixels as small, 144 to 1024 pixels as medium, and 1024 and above as large. The percentage of small, medium and large objects in iSAID is 52.0, 33.7 and 9.7, respectively. The box plot in Fig.~\ref{fig:mean_area_per_category} presents statistics of area for each class of iSAID. It can be seen that the size of objects varies greatly both among and across classes. For instance, the ship category contains small boats covering area of $10$ pixels, as well as, large vessels of sizes upto 1,436,401 pixels, depicting a huge intra-class variation. Similarly, a small vehicle can be as small as 10 pixels and a ground track field can be as large as 1,297,121 pixels, illustrating immense inter-class variation. Fig.~\ref{fig:scale variation} shows the variation in scale when small and large objects of same or different categories appear together, which is a very common case in aerial imagery. We can notice that the ratio between the area of the largest object and the smallest object can reach up to 20,000.  Such enormous scale variation poses an extreme challenge for instance segmentation methods that need to handle both tiny and very large objects, simultaneously.

\noindent
\textbf{Aspect ratio.}
In aerial images many objects occur with unusually large aspect ratios, which is not the case in traditional ground images. Fig.~\ref{fig:aspect ratio} depicts the distribution of aspect ratio for object instances in our proposed dataset. We can notice that instances exhibit huge variation in aspect ratios, reaching up to 90 (with an average of 2.4). Moreover, a large number of instances present in our dataset have a large aspect ratio. %Instances of roundabout and ground track field classes are of small aspect ratio, whereas bridge and harbour have a large aspect ratio.    

\begin{table*}[t]
  \small
\begin{minipage}{0.49\textwidth}
\centering
\setlength{\tabcolsep}{3.2pt}
    \begin{tabular}{l|ccc|ccc}
      Method & AP & $AP_{50}$ & $AP_{75}$ & $AP_{S}$ & $AP_{M}$ & $AP_{L}$  \\ %
      \hline
      Mask R-CNN \cite{he2017mask} & 25.65&51.30&22.72&14.46&31.26&37.71 \\ % <--
      {Mask R-CNN+} & 33.41&56.77&34.66&35.83&46.50&23.93\\
      \hline
      PANet \cite{liu2018path} &   34.17&56.57&35.84&19.56&42.27&\textbf{46.62} \\
      PANet+  &  39.54&63.59&42.22&42.14&53.61&38.50 \\
      \hline
      %Mask R-CNN++ &  26.30&50.00&25.70&28.30&37.00&12.20\\
      PANet++ & \textbf{40.00} & \textbf{64.54} & \textbf{42.50} & \textbf{42.46} & \textbf{54.74} & 43.16 
      %\hline
    \end{tabular}\vspace{-0.2cm}
    \captionof{table}{\textbf{Instance segmentation} results using mask AP on iSAID test set. PANet \cite{liu2018path} and its variants outperform Mask R-CNN \cite{he2017mask} and its variants with significant margin. PANet++ with backbone ResNet-152 performs best. %Improved results are obtained by making minor modifications in baseline methods, as shown in the rows of Mask R-CNN+, PANet+ and PANet++.
    }
  \label{tab:res_ins_seg}\vspace{-0.1cm}
\end{minipage}
\hfill
\begin{minipage}{0.49\textwidth}
\centering
    \setlength{\tabcolsep}{1.8pt}
    \begin{tabular}{l|ccc|ccc}
      Method &  $AP^{bb}$ & $AP^{bb}_{50}$ & $AP^{bb}_{75}$ & $AP^{bb}_{S}$ & $AP^{bb}_{M}$ & $AP^{bb}_{L}$  \\ %
      \hline
      Mask R-CNN \cite{he2017mask} &  36.50&59.06&41.27&26.16&43.10&43.32\\ % <--
     {Mask R-CNN+} &  37.18&60.79&40.67&39.84&43.72&16.01\\
      \hline
      PANet \cite{liu2018path} &   41.66&60.94&46.62&26.92&47.81&\textbf{50.95}\\
      {PANet+} &  46.31&66.90&51.68&48.92&53.33&26.52\\
      \hline
      %Mask R-CNN++ &   31.10&51.30&33.50&33.50&35.40&8.50\\
      PANet++ & \textbf{47.0}& \textbf{68.06} & \textbf{52.37} & \textbf{49.48} & \textbf{55.07} & 27.97
      %\hline
    \end{tabular}\vspace{-0.2cm}
    \captionof{table}{\textbf{Object detection} results using bounding box AP on iSAID test set. Similar to instance segmentation case, PANet \cite{liu2018path} and its variants generate better results than Mask-RCNN and its variants. 
    %The variant of PANet with minor modifications (PANet++) leads to the best performance. 
    }
  \label{tab:res_bb_det}\vspace{-0.1cm}
\end{minipage}
\end{table*}

\begin{table*}[t]
  \small
    \begin{center}
    \setlength{\tabcolsep}{3.3pt}
    \begin{tabular}{l|cc|ccccccccccccccc}
       Method &  AP & $AP_{50}$ & Plane & BD & Bridge & GTF & SV & LV & Ship & TC & BC & ST & SBF & RA & Harbor & SP & HC   \\ %
      \hline
      Mask R-CNN \cite{he2017mask} & 25.7&51.3&37.7&42.5&13.0&23.6&6.9&7.4&26.6&54.9&34.6&28.0&20.8&35.9&22.5&25.1&5.3\\ % <--
      {Mask R-CNN+} & 33.4&56.8&41.7&39.6&15.2&25.9&16.9&30.4&48.8&72.9&43.1&32.0&26.7&36.0&29.6&36.7&5.6\\
      %{Mask R-CNN++} &  26.3&50.0&33.6&34.6&13.0&22.0&13.0&20.2&41.0&67.1&30.0&32.0&19.0&	12.0&20.0&31.3&10.0 \\
      \hline
      PANet & 34.2&56.8&39.2&45.5&15.1&29.3&15.0&28.8&45.9&74.1&47.4&29.6&33.9&36.9	&26.3&36.1&9.5\\
      %\hline
      PANet++ & \textbf{40.0}& \textbf{64.6}& \textbf{48.7}& \textbf{50.3}& \textbf{18.9}&\textbf{32.5}& \textbf{20.4} & \textbf{34.4} & \textbf{56.5} & \textbf{78.4} & \textbf{52.3} & \textbf{35.4} & \textbf{38.8} & \textbf{40.2} & \textbf{35.8} & \textbf{42.5} & \textbf{13.7}
      %\hline
    \end{tabular}\vspace{-0.5cm}
    \end{center}
     \caption{\textbf{Class-wise instance segmentation} results on iSAID test set. Note that short names are used to define categories: BD-Baseball diamond, GTF-Ground field track, SV-Small vehicle, LV-Large vehicle TC-Tennis court, BC-Basketball court, SC-Storage tank, SBF-Soccer-ball field, RA-Roundabout, SP-Swimming pool, and HC-Helicopter.}
     \label{tab:class_wise}\vspace{-0.1cm}
\end{table*}

\begin{table*}[t]
  \small
  \begin{center}
    \setlength{\tabcolsep}{3.4pt}
    \begin{tabular}{l|cc|ccccccccccccccc}
       Method & $AP^{bb}$ & $AP^{bb}_{50}$ & Plane & BD & Bridge & GTF & SV & LV & Ship & TC & BC & ST & SBF & RA & Harbor & SP & HC   \\ %
      \hline
      Mask R-CNN \cite{he2017mask} &  36.6&59.1&57.8&44.7&19.7&36.4&17.9&31.7&	46.9&70.2&42.7&31.4&25.4&36.4&41.0&36.2&21.9\\ % <--
      %\hline
      {Mask R-CNN+} & 37.2&60.8&58.5&38.5&18.6&32.7&20.8&36.8&51.4&72.9&43.1&32.0&	26.7&36.0&29.6&48.8&29.6\\
      %{Mask R-CNN++} &31.1&51.3&55.1&33.4&16.3&28.8&16.5&28.8&44.9&68&32.2&31.9&19.1&	12.3&33.5&35.5&10.0\\
      \hline
      PANet  &  41.7&61.0&62.8&47.5&19.3&44.3&18.3&35.0&50.3&77.4&48.5&30.9&35.3&40.4&46.6&40.4&27.9\\
      
      PANet++ &  \textbf{47.0} & \textbf{68.1} & \textbf{68.1} & \textbf{51.0} & \textbf{23.4} & \textbf{44.2} & \textbf{27.3} & \textbf{42.1} &\textbf{61.9} & \textbf{79.4}& \textbf{53.8} & \textbf{38.1} & \textbf{39.1}& \textbf{43.4}& \textbf{53.6} & \textbf{47.1} & \textbf{32.4}
      %\hline
    \end{tabular}\vspace{-0.5cm}
    \end{center}
  \caption{\textbf{Class-wise object detection} results on iSAID test set. The same short names for categories are used as in Table~\ref{tab:class_wise}.
  }
  \label{tab:bb_res_cl}\vspace{-0.1cm}
\end{table*}

%\vspace{-0.9em}
\section{Experiments}
In this section, we test how general instance segmentation methods, particularly developed for regular scene datasets, perform on our newly developed aerial dataset (some sample images are shown in Fig. \ref{fig:our samples}).  To this end, we use MaskR-CNN~\cite{he2017mask} and PANet~\cite{liu2018path}: the former for its popularity as a meta algorithm and the latter for its state-of-the-art results. 
Additionally, we make simple modifications in the baseline models and report the results of these variants. For evaluation, we use the standard COCO metrics: AP (averaged over IoU threshold), AP$_{50}$, AP$_{75}$, AP$_{S}$, AP$_{M}$ and AP$_{L}$, where $S$, $M$ and $L$ represent small (area: 10-144 pixels), medium (area:144 to 1024 pixels) and large objects (area:1024 and above), respectively.

\begin{figure*}[t] %fig7
\begin{center}
    \begin{subfigure}[t]{0.19\textwidth} %fig10
      \includegraphics[width=\linewidth]{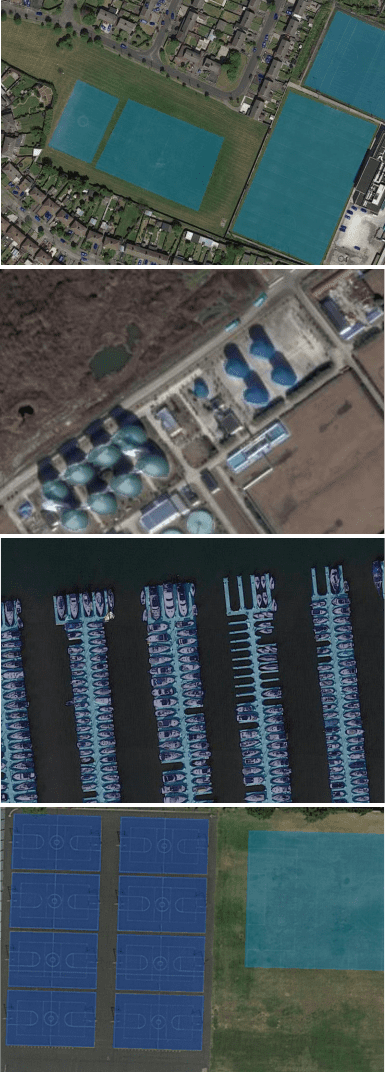}
       \caption{Ground Truth}
      \label{}
    \end{subfigure}
    \begin{subfigure}[t]{0.19\textwidth} %fig10
      \includegraphics[width=\linewidth]{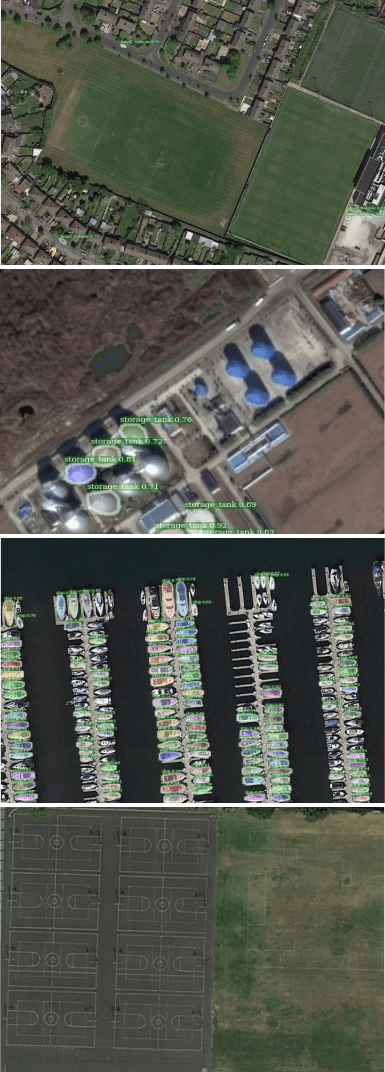}
       \caption{Mask R-CNN \cite{he2017mask}}
      \label{}
    \end{subfigure}
    \begin{subfigure}[t]{0.19\textwidth} %fig10
      \includegraphics[width=\linewidth]{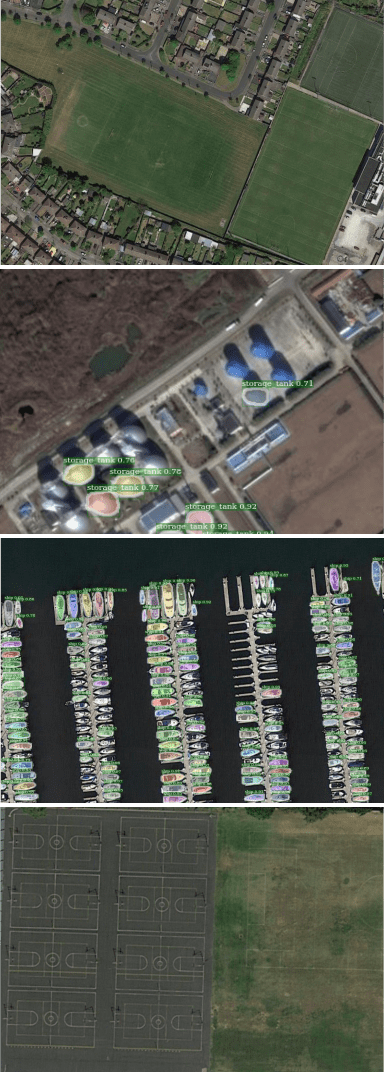}
       \caption{Mask R-CNN+}
      \label{}
    \end{subfigure}
    \begin{subfigure}[t]{0.19\textwidth} %fig10
      \includegraphics[width=\linewidth]{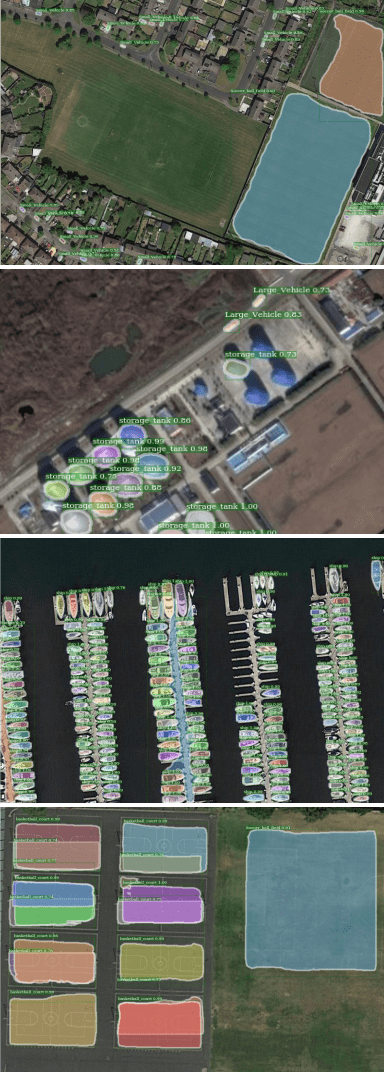}
       \caption{PANet \cite{liu2018path}}
      \label{}
    \end{subfigure}
    \begin{subfigure}[t]{0.19\textwidth}%fig10
      \includegraphics[width=\linewidth]{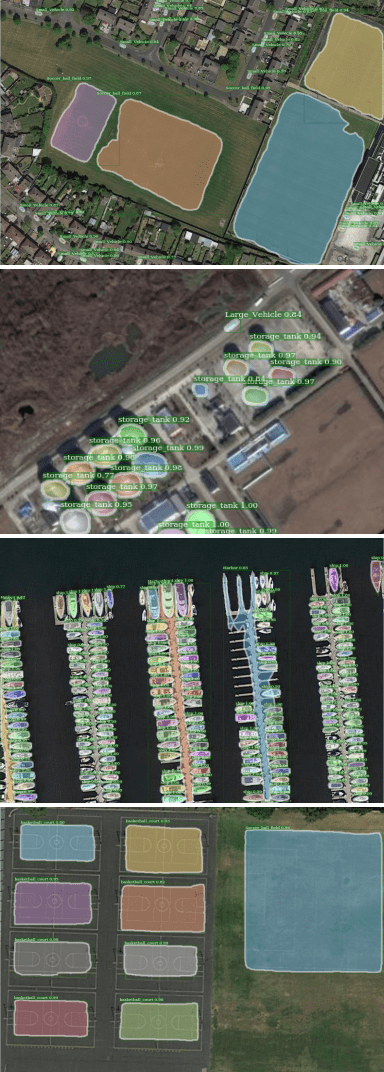}
       \caption{PANet++}
      \label{}
    \end{subfigure}
\end{center}\vspace{-1.2em}
\caption{Visual results on images from test set of iSAID. It can be noticed that the original Mask R-CNN \cite{he2017mask} yields the least accurate results, with missing object instances. Whereas, PANet++ produces significantly better results compared to its original counter part \cite{liu2018path}, as well as Mask R-CNN and Mask R-CNN+.}
\label{fig:qual_res}%\vspace{-0.2cm}
\end{figure*}

\noindent 
\textbf{Implementation Details.} Images with large resolution (\eg 4000 pixels in width) are commonly present in iSAID. The baseline methods \cite{he2017mask,liu2018path} cannot handle images with such unusually large spatial dimension. Therefore, we opt to train and test the baseline methods on the patches of size $800{\times}800$ extracted from the full resolution images with a stride set to 200. In order to train baseline Mask R-CNN and PANet models, we use the same hyper-parameters as in the original papers \cite{he2017mask,liu2018path}.
In the training phase, the cropped patches are re-scaled with shorter edges as 800 pixels and longer edges as 1400 pixels. During the cropping process, some objects may get cut. we then generate new annotations for the patches with updated segmentation masks.
We use mini-batch size of 16 for training. Our models are trained on 8 GPUs for 180k iterations with an initial learning rate of 0.025, that is decreased by a factor of 10 at 90k iteration. We use weight decay of 0.0001 and momentum of 0.9.

 In an effort to benchmark the proposed dataset, we consider the original Mask R-CNN \cite{he2017mask} and PANet \cite{liu2018path} as our baseline models, both using ResNet101-FPN as backbone. We do not change any hyper-parameter settings in the baseline models. On top of these baselines, we make three minor modifications to develop Mask R-CNN+ and PANet+: \textbf{(a)} Since, large number of objects are present per image, we consider the number of detection boxes to be 1000 (instead of 100 considered by default in the baselines) during evaluation. \textbf{(b)} As high scale variation exists within aerial images, we use scale augmentations at six scales (1200,1000,800,600,400). In comparison, the baseline considers a single scale of 800 pixels (shorter side). \textbf{(c)} An NMS (non-maximal suppression) threshold of 0.6 is used instead of the 0.5 used for baseline. %The models obtained after the above three modifications are denoted by Mask R-CNN+ and PANet+, respectively. 
Lastly, for our best model (PANet), we also try a heavier backbone (ResNet-152-FPN) that results in the top performing models for instance segmentation and bounding box detection. We term this model as PANet++. Note that the modifications in baselines are minor, and we expect that more sophisticated algorithmic choices might significantly improve the results.

% \textbf{Inference}: 
% Similar to train stage, the baseline methods are applied at inference time on the cropped patches from the test images to provide us with %temporary 
% instance segmentation results, on which the NMS is applied. 
%For Baseline methods, the mask branch is applied to the highest scoring 100 detection boxes. For our method we changed the number of detection boxes to 2000 because of the increase in number of objects.

% \iffalse
% \textcolor{blue}{Akshita: please write this section (here are some questions that you can use as guideline while writing):
% \begin{itemize}
%     \item Are you using default parameters for baseline methods (Mask R-CNN and PANet?, if yes, then just refer the reader to find these parameters in the corresponding papers.)
%     \item What are modifications that you made in baselines? (explain them here.)
%     %panet heavier head resnet 152 and scale augmentations (640, 672, 736, 768, 800, 930)
%     %maskrcnn heavier head resnet 152 and scale augmentations 
%     %inference boundin box 2000
% \end{itemize}
% }
% \fi

\subsection{Results}
%So we use ResneXt-101-FPN as backbone with scale augmentations  that improved the performance from 32.24\% to 36.43\% and denoted the model as PANet+. We applied the same augmentations to Mask R-CNN with ResneXt-101-FPN backbone denoted as Mask R-CNN+.  we experimentally found that replacing the current backbone for PANet with a heavier one i.e, ResneXt-101-FPN is helpful in attaining better performance. 
In Table~\ref{tab:res_ins_seg}, we report the results achieved by baselines (Mask R-CNN \cite{he2017mask} and PANet \cite{liu2018path}) and their variants for the instance segmentation task. It can be seen that the PANet \cite{liu2018path} with its default parameters outperforms the Mask R-CNN \cite{he2017mask} on iSAID. This trend is similar to the performance of these baselines on the MSCOCO dataset for instance segmentation in regular ground images. Moreover, by making minor modifications in baselines to make them suitable for aerial images, we were able to obtain marginal improvements e.g., an absolute increment of 7.8 AP with Mask R-CNN+ over baseline \cite{he2017mask}. The best performance is achieved by PANet++ which uses a stronger ResNet-152-FPN backbone. To study the performance trend for different classes, we also report class-wise AP in Table~\ref{tab:class_wise}. Notably, in the case of PANet++, we observe a significant performance gain of $\sim$5 points or more in AP$_{50}$ for some categories such as baseball diamond, basketball court and harbour. %Whereas for Mask R-CNN, the performance is marginally improved, indicating the need of more sophisticated design choices.

In addition to instance segmentation masks, we also compute bounding-box object detection results, as reported in Tables \ref{tab:res_bb_det} and \ref{tab:bb_res_cl}. In this experiment, the horizontal bounding-boxes are considered. For object detection, we observe similar trends in methods' ranking as they were for instance segmentation. It is important to note that our results are inferior to those reported in \cite{dota}, possibly due to the large number of newly introduced object instances in iSAID (655,451 vs 188,282 in DOTA).

Qualitative results for instance segmentation are shown in Fig.~\ref{fig:qual_res}. The results are shown for Mask R-CNN and PANet baselines and their modified versions. We note that with simple modifcations to these strong baselines, we were able to significantly improve on extreme sized objects (both very small and large objects). As expected from the quantitative results, the PANet++ achieves most convincing qualitative results with accurate instance masks among the other evaluated models.

% \iffalse
% \begin{table*}[]
% \centering
% \setlength{\tabcolsep}{5pt}
% \begin{tabular}{c|c|c|c|c}
% \hline
%                   & Mask R-CNN & PANet & Mask R-CNN+ & PANet+  \\
% \hline
% Ship               & 0        & 0     & 0         & 0 \\
% Storage Tank       & 0        & 0     & 0         & 0 \\
% Baseball Diamond   & 0        & 0     & 0         & 0 \\
% Tennis Court       & 0        & 0     & 0         & 0 \\
% Basketball Court   & 0        & 0     & 0         & 0 \\
% Ground Track Field & 0        & 0     & 0         & 0 \\
% Bridge             & 0        & 0     & 0         & 0 \\
% Large Vehicle      & 0        & 0     & 0         & 0 \\
% Small Vehicle      & 0        & 0     & 0         & 0 \\
% Helicopter         & 0        & 0     & 0         & 0 \\
% Swimming Pool      & 0        & 0     & 0         & 0 \\
% Roundabout         & 0        & 0     & 0         & 0 \\
% Soccer Ball Field  & 0        & 0     & 0         & 0 \\
% Plane              & 0        & 0     & 0         & 0 \\
% Harbor             & 0        & 0     & 0         & 0 \\
% \hline
% Avg.                & 0        & 0     & 0         & 0 \\
% \hline
% \end{tabular}
% \caption{Benchmark results on the methods.}
% \label{Table:benchmark_results}
% \end{table*}
% \fi

\section{Conclusion}\vspace{-0.2em}
Delineating each object instance in aerial images is a practically significant and a scientifically challenging problem. The progress in this area has been limited due to the lack of large-scale, densely annotated satellite image dataset with accurate instance masks. To bridge this gap, we propose a new instance segmentation dataset which encompasses 15 object categories and 655,451 instances in total. We extensively benchmark the dataset on instance segmentation and object detection tasks. Our results show that the aerial imagery pose new challenges to existing instance segmentation algorithms such as a large number of objects per image, limited appearance details, several small objects, significant scale variations among the different object types and a high class imbalance. We hope that our contribution will lead to new developments on the instance segmentation task in aerial imagery.

{\small
\bibliographystyle{ieee}
\bibliography{egbib}

\begin{thebibliography}{10}\itemsep=-1pt

\bibitem{airbus}
Dataset for airbus ship dectection challenge.
\newblock \url{https://www.kaggle.com/c/airbus-ship-detection/}, 2018.
\newblock [Online; accessed 27-May-2019].

\bibitem{Anwer2018}
R.~M. Anwer, F.~S. Khan, J.~van~de Weijer, M.~Molinier, and J.~Laaksonen.
\newblock Binary patterns encoded convolutional neural networks for texture
  recognition and remote sensing scene classification.
\newblock {\em ISPRS journal of photogrammetry and remote sensing}, 138:74--85,
  2018.

\bibitem{sztaki}
C.~Benedek, X.~Descombes, and J.~Zerubia.
\newblock Building development monitoring in multitemporal remotely sensed
  image pairs with stochastic birth-death dynamics.
\newblock {\em TPAMI}, 34(1):33--50, 2012.

\bibitem{chen2018encoder}
L.-C. Chen, Y.~Zhu, G.~Papandreou, F.~Schroff, and H.~Adam.
\newblock Encoder-decoder with atrous separable convolution for semantic image
  segmentation.
\newblock In {\em ECCV}, 2018.

\bibitem{nwpu}
G.~{Cheng}, P.~{Zhou}, and J.~{Han}.
\newblock Learning rotation-invariant convolutional neural networks for object
  detection in {VHR} optical remote sensing images.
\newblock {\em IEEE Transactions on Geoscience and Remote Sensing},
  54(12):7405--7415, 2016.

\bibitem{cityscapes}
M.~Cordts, M.~Omran, S.~Ramos, T.~Rehfeld, M.~Enzweiler, R.~Benenson,
  U.~Franke, S.~Roth, and B.~Schiele.
\newblock The cityscapes dataset for semantic urban scene understanding.
\newblock In {\em CVPR}, 2016.

\bibitem{imagenet}
J.~Deng, W.~Dong, R.~Socher, L.-J. Li, K.~Li, and L.~Fei-Fei.
\newblock {ImageNet}: A large-scale hierarchical image database.
\newblock In {\em CVPR}, 2009.

\bibitem{pascal}
M.~Everingham, L.~Van~Gool, C.~K. Williams, J.~Winn, and A.~Zisserman.
\newblock The pascal visual object classes (voc) challenge.
\newblock {\em IJCV}, 88(2):303--338, 2010.

\bibitem{Urban3D2017}
H.~Goldberg, M.~Brown, and S.~Wang.
\newblock A benchmark for building footprint classification using
  orthorectified {RGB} imagery and digital surface models from commercial
  satellites.
\newblock In {\em Proceedings of IEEE Applied Imagery Pattern Recognition
  Workshop}, 2017.

\bibitem{he2017mask}
K.~He, G.~Gkioxari, P.~Doll{\'a}r, and R.~Girshick.
\newblock Mask r-cnn.
\newblock In {\em ICCV}, 2017.

\bibitem{He2016}
K.~He, X.~Zhang, S.~Ren, and J.~Sun.
\newblock Deep residual learning for image recognition.
\newblock In {\em CVPR}, 2016.

\bibitem{tas}
G.~Heitz and D.~Koller.
\newblock Learning spatial context: Using stuff to find things.
\newblock In {\em EECV}, 2008.

\bibitem{khan2019striking}
S.~Khan, M.~Hayat, S.~W. Zamir, J.~Shen, and L.~Shao.
\newblock Striking the right balance with uncertainty.
\newblock 2019.

\bibitem{khan2018guide}
S.~Khan, H.~Rahmani, S.~A.~A. Shah, and M.~Bennamoun.
\newblock A guide to convolutional neural networks for computer vision.
\newblock {\em Synthesis Lectures on Computer Vision}, 8(1):1--207, 2018.

\bibitem{khan2017forest}
S.~H. Khan, X.~He, F.~Porikli, and M.~Bennamoun.
\newblock Forest change detection in incomplete satellite images with deep
  neural networks.
\newblock {\em IEEE Transactions on Geoscience and Remote Sensing},
  55(9):5407--5423, 2017.

\bibitem{xview}
D.~Lam, R.~Kuzma, K.~McGee, S.~Dooley, M.~Laielli, M.~Klaric, Y.~Bulatov, and
  B.~McCord.
\newblock {xView}: Objects in context in overhead imagery.
\newblock In {\em arXiv}, 2018.

\bibitem{mscoco}
T.-Y. Lin, M.~Maire, S.~Belongie, J.~Hays, P.~Perona, D.~Ramanan,
  P.~Doll{\'a}r, and C.~L. Zitnick.
\newblock Microsoft {COCO}: Common objects in context.
\newblock In {\em ECCV}, 2014.

\bibitem{liu2018path}
S.~Liu, L.~Qi, H.~Qin, J.~Shi, and J.~Jia.
\newblock Path aggregation network for instance segmentation.
\newblock In {\em CVPR}, 2018.

\bibitem{liu2016ssd}
W.~Liu, D.~Anguelov, D.~Erhan, C.~Szegedy, S.~Reed, C.-Y. Fu, and A.~C. Berg.
\newblock {SSD}: Single shot multibox detector.
\newblock In {\em ECCV}, 2016.

\bibitem{hrsc}
Z.~Liu, H.~Wang, L.~Weng, and Y.~Yang.
\newblock Ship rotated bounding box space for ship extraction from
  high-resolution optical satellite images with complex backgrounds.
\newblock {\em IEEE Geoscience and Remote Sensing Letters}, 13(8):1074--1078,
  2016.

\bibitem{maggiori2017dataset}
E.~Maggiori, Y.~Tarabalka, G.~Charpiat, and P.~Alliez.
\newblock Can semantic labeling methods generalize to any city? the inria
  aerial image labeling benchmark.
\newblock In {\em IEEE International Geoscience and Remote Sensing Symposium
  (IGARSS)}, 2017.

\bibitem{marcos2018land}
D.~Marcos, M.~Volpi, B.~Kellenberger, and D.~Tuia.
\newblock Land cover mapping at very high resolution with rotation equivariant
  cnns: Towards small yet accurate models.
\newblock {\em ISPRS Journal of Photogrammetry and Remote Sensing},
  145:96--107, 2018.

\bibitem{cowc}
T.~N. Mundhenk, G.~Konjevod, W.~A. Sakla, and K.~Boakye.
\newblock A large contextual dataset for classification, detection and counting
  of cars with deep learning.
\newblock In {\em ECCV}, 2016.

\bibitem{nyu}
P.~K. Nathan~Silberman, Derek~Hoiem and R.~Fergus.
\newblock Indoor segmentation and support inference from {RGBD} images.
\newblock In {\em ECCV}, 2012.

\bibitem{vedai}
S.~Razakarivony and F.~Jurie.
\newblock Vehicle detection in aerial imagery: A small target detection
  benchmark.
\newblock {\em Journal of Visual Communication and Image Representation},
  34:187--203, 2016.

\bibitem{redmon2017yolo9000}
J.~Redmon and A.~Farhadi.
\newblock {YOLO9000:} better, faster, stronger.
\newblock In {\em CVPR}, 2017.

\bibitem{ren2015faster}
S.~Ren, K.~He, R.~Girshick, and J.~Sun.
\newblock Faster {R-CNN}: Towards real-time object detection with region
  proposal networks.
\newblock In {\em NIPs}, 2015.

\bibitem{santamaria2017mass}
C.~Santamaria, M.~Alvarez, H.~Greidanus, V.~Syrris, P.~Soille, and
  P.~Argentieri.
\newblock Mass processing of sentinel-1 images for maritime surveillance.
\newblock {\em Remote Sensing}, 9(7):678, 2017.

\bibitem{Simonyan2014}
K.~Simonyan and A.~Zisserman.
\newblock Very deep convolutional networks for large-scale image recognition.
\newblock {\em ICLR}, 2014.

\bibitem{szegedy2017inception}
C.~Szegedy, S.~Ioffe, V.~Vanhoucke, and A.~A. Alemi.
\newblock Inception-v4, inception-{ResNet} and the impact of residual
  connections on learning.
\newblock In {\em AAAI}, 2017.

\bibitem{spacenet}
N.~Weir, D.~Lindenbaum, A.~Bastidas, A.~Van~Etten, S.~McPherson, J.~Shermeyer,
  V.~Kumar, and H.~Tang.
\newblock {SpaceNet MVOI:} a multi-view overhead imagery dataset.
\newblock {\em arXiv}, 2019.

\bibitem{dota}
G.-S. Xia, X.~Bai, J.~Ding, Z.~Zhu, S.~Belongie, J.~Luo, M.~Datcu, M.~Pelillo,
  and L.~Zhang.
\newblock {DOTA}: A large-scale dataset for object detection in aerial images.
\newblock In {\em CVPR}, 2018.

\bibitem{places}
B.~Zhou, A.~Lapedriza, J.~Xiao, A.~Torralba, and A.~Oliva.
\newblock Learning deep features for scene recognition using places database.
\newblock In {\em NIPS}, 2014.

\bibitem{ade20k}
B.~Zhou, H.~Zhao, X.~Puig, S.~Fidler, A.~Barriuso, and A.~Torralba.
\newblock Scene parsing through ade20k dataset.
\newblock In {\em CVPR}, 2017.

\bibitem{ucas}
H.~Zhu, X.~Chen, W.~Dai, K.~Fu, Q.~Ye, and J.~Jiao.
\newblock Orientation robust object detection in aerial images using deep
  convolutional neural network.
\newblock In {\em ICIP}, 2015.

\end{thebibliography}
}

\end{document}